# Towards retrieving dispersion profiles using quantum-mimic Optical Coherence Tomography and Machine Learning

Krzysztof A. Maliszewski,[1,†] Piotr Kolenderski,[2] Varvara Vetrova,[1] Sylwia M. Kolenderska[3,*,†]

[1] *School of Mathematics and Statistics, University of Canterbury, Christchurch, New Zealand*

[2] *Faculty of Physics, Astronomy and Informatics, Nicolaus Copernicus University, Toruń, Poland*

[3] *School of Physical and Chemical Sciences, University of Canterbury, Christchurch, New Zealand*

[†] *These authors contributed equally to this work*

[*]*skol745@aucklanduni.ac.nz*

**Abstract:** Artefacts in quantum-mimic Optical Coherence Tomography are considered detrimental because they scramble the images even for the simplest objects. They are a side effect of autocorrelation which is used in the quantum entanglement mimicking algorithm behind this method. Interestingly, the autocorrelation imprints certain characteristics onto an artefact - it makes its shape and characteristics depend on the amount of dispersion exhibited by the layer that artefact corresponds to. This unique relationship between the artefact and the layer's dispersion can be used to determine Group Velocity Dispersion (GVD) values of object layers and, based on them, build a dispersion-contrasted depth profile. The retrieval of GVD profiles is achieved via Machine Learning. During training, a neural network learns the relationship between GVD and the artefacts' shape and characteristics, and consequently, it is able to provide a good qualitative representation of object's dispersion profile for never-seen-before data: computer-generated single dispersive layers and experimental pieces of glass.

**Introduction**

Optical Coherence Tomography (OCT) is a non-contact and non-invasive light-based method for obtaining high-resolution three-dimensional images of objects semi-transparent to light, such us the eye or skin but also fruits (kiwis, blueberries) or synthetic materials [1]. The core of OCT is an interferometer where the light backscattered from the layers of the object in the object arm and the light reflected from the mirror in the reference arm interfere. In the most popular type of OCT, Fourier domain OCT, a spectrum is acquired at the output of the interferometer. Since the light interference causes fringes to appear in the detected spectrum, one applies Fourier transformation to retrieve the object's depth profile called an A-scan, i.e. a one-dimensional signal representing the internal structure of the object.

Wavelength-dependent variations of the refractive index in an object leads to the phenomenon of chromatic dispersion. Due to dispersion, different "wavelengths" travel at different speeds and consequently, arrive at different times at the detector. In OCT, chromatic dispersion is considered with regards to its relative amounts in the arms of the interferometer. If the dispersion is balanced - so the amount of dispersion in the object arm is the same as in the reference arm - the light in the object arm and the light in the reference arm will experience the same time delays and arrive at the detector to create uniform fringes. If the chromatic dispersion is unbalanced, the resultant fringes will not be uniform. Their distribution will be nonlinear and the Fourier transform of such fringes will result in a broadened peak. Since the axial resolution in OCT is defined as full width at half maximum (FWHM) of a peak in the A-scan, the unbalanced dispersion will effectively be responsible for resolution degradation. The interferometer dispersion imbalance is compensated using either hardware [2] or software methods [3].

Since the dispersion is usually different for a different layer of the imaged object, it is not possible to match all of them at the same time. As a result, traces of dispersion-related peak broadening are left, especially for deeper layers due to the dispersion's accumulative nature. However, this behaviour can be used to one's advantage: it was shown that layer's second order of dispersion, Group Velocity Dispersion (GVD), values can be extracted to characterise the imaged object [4–6] or even to be correlated with early signs or progression of diseases [7]. Unfortunately, the current methods for GVD values extraction are very error-prone [7] or work only for very simple objects [4,5].

Quantum-mimic OCT provides very favourable enhancements to OCT imaging: resolution increase and even orders dispersion cancellation [8,9]. At the same time, it generates artefacts – additional peaks which do not correspond to the structure of the imaged object. A large number of such artefacts lead to the images of even the simplest of objects being scrambled. This is why the artefacts have always been regarded as a detrimental side effect and efforts were put into finding methods for removing them [8]. However, the artefacts contain a lot of object-related information which is encoded in their location and behaviour. More specifically, the artefact shape changes in the presence of nonlinearities in the spectrum. This means that the artefacts are sensitive to any nonlinearity-inducing phenomena, especially chromatic dispersion. Since an object layer generates a set of artefacts whose location in the image is unique, the dispersion-originating behaviour could in principle be analysed for each layer separately to determine its GVD value. Because for multi-layered objects, it is impossible to isolate a specific layer and its artefacts using standard data-processing approaches, a Machine Learning solution is employed.

In our work, we use the quantum-mimic OCT modality called Intensity Correlation OCT (ICA-OCT) [8], where the signal is generated by applying a simple algorithmic procedure to a raw OCT spectrum. We show that a neural network trained on perfect, synthetic ICA-OCT signals is able to near-perfectly predict a dispersion profile - a distribution of GVD values within an A-scan - for simple computer-generated objects as well as for the experimental data representing pieces of glass: quartz, BK7 and sapphire. We also provide an analysis of our neural network's performance with regards to the presence of autocorrelation peaks, i.e. peaks which come from the interference of light back-scattered from the object itself.

**Methods**

*Data*

Our dataset consists of synthetically generated objects with a random number of interfaces, up-to 12, placed at random locations in an A-scan. The data did not incorporate noise. We use FFT stacks as inputs and corresponding dispersion profiles as labels (output data). FFT stacks are created using the algorithm in Ref. [9]: first a spectrum is synthesised (1024 element-long, centred at 840 nm, and with the total spectral bandwidth of 160 nm and a Gaussian profile), then split into 50 fragments which are autocorrelated, zero-padded to be 2048 element-long and Fourier transformed. Only half of a Fourier transform is taken, which means that the resultant FFT stack size is 50 by 1024 elements.

The dispersion profiles represent the GVD value distribution within an object. They are 1024-element-long vectors whose elements are in the range [0,1] corresponding to the GVD value range of (-5000, 5000) fs$^2$/mm.

Our training dataset consisted of 260,000 examples with an average of around 24,000 stack/dispersion profile pairs for objects with 2 to 12 interfaces and 512 pairs for objects with one interface. The validation dataset contained around 20,000 stack/dispersion profile pairs with an average of 2,000 examples per object type (the type being the number of interfaces) excluding any single-interface examples.

*Neural network*

We based our machine learning model on a modified VGG-16 architecture. We found optimum neural network hyper-parameters using an automatic optimization software framework called Optuna. The list of hyper-parameters that underwent optimisation is presented in the centre column in Table 1.

The search for optimum hyper-parameters was carried out using a dataset comprising a total of 15,000 examples. The goal of the optimization was to minimize the loss function. Model's performance was evaluated after 10 epochs. The hyper-parameters for which the minimum loss was obtained are presented in the right-most column in Table 1.

Table 1. List of optimized hyper-parameters and their optimal values. [a..b, c] - a and b are the limits of the range, c is the step size.

| Hyper-parameter | Values | Optimal Value |
| --- | --- | --- |
| Batch norm after Conv2D | True, False | True |
| Pooling type | max, avg | max |
| Number of fully connected layers | [1..4, 1] | 1 |
| Number of units in each fully connected layer | [1024..16384, 1024] | 14336 |
| Fully connected layer normalization | Batch normalization, Layer normalization, No normalization | Layer normalization |
| Dropout rates | [0.1..0.5, 0.05] | 0.1 |
| Loss function | Binary cross-entropy, Mean Absolute Error, Mean Squared Error, Custom loss functions | Mean Absolute Error |
| Learning rate | [0.1..1e-07, a variable step size] | 0.0001 |
| Optimizers | Adam, RMSProp, SGD | Adam |

The schematics of the final, most optimum model architecture is presented in Fig. 1.

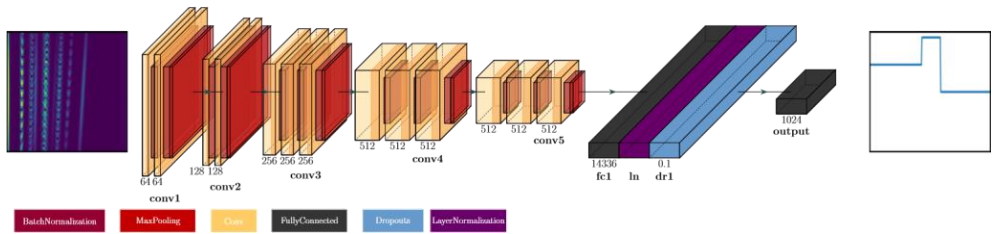

Fig. 1. Schematics of the optimised neural network architecture.

The training was performed on a computer with NVIDIA GeForce GTX 3060 12GB graphics card. The batch size was set to 16 and, since the data was generated on the fly, each epoch took around 1.5 hours to finish. We trained our model for 100 epochs.

*OCT system*

The experimental data was acquired using an OCT system where the light source is a Superluminescent Diode with the central wavelength of 840 nm and the total spectral bandwidth of 160 nm, the interferometer contains a 50:50 beamsplitter and a 50-mm focal-length achromatic lens in each arm, and the spectrometer is an Optical Spectrum Analyser which outputs 2048-point long spectra with the spectral resolution of 0.1 nm. The axial resolution in air is 4.1 µm and the 6-dB fall-off 1.4 mm.

**Results**

The trained neural network is first tested on computer-generated data representing a two-interface object, i.e. a single layer. Its performance is analysed in terms of the position of an autocorrelation peak in the A-scan - with the theoretical background supporting the results provided in Supplementary Document - and with regards to noise. Next, the same neural network is applied to experimental data corresponding to quartz, BK7 and sapphire glasses.

*Dispersion profiles - computer-generated data for two-interface objects*

An FFT stack is generated for a two-interface object with a layer dispersion of 2,500 $fs^2$/mm with no noise and no autocorrelation peak (Fig. 2a). The resulting dispersion profile prediction (Fig. 2b, orange line) remains in good agreement with the ground truth (blue line), except for the area close to 0 optical path difference (OPD, 0 OPD is effectively the beginning of the X axis).

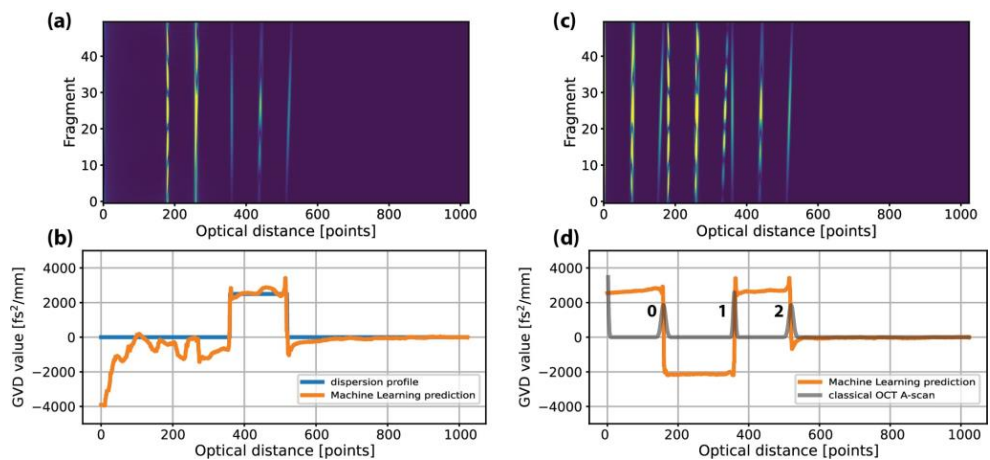

Fig. 2. A two-interface object with object dispersion equal to 2,500 $fs^2$/mm. **(a)** FFT stack - an input of the neural network - does not contain an autocorrelation peak. **(b)** The neural network prediction - orange line - is in good agreement with the ground truth - blue line. When **(c)** the FFT stack contains the autocorrelation peak, **(d)** the neural network prediction - orange line - shows a dispersion profile affected around the location of the autocorrelation peak (marked as 0 on an A-scan plotted in gray), which is positioned in front of the object (object peaks marked as 1 and 2).

When the same FFT stack incorporates the contribution of an autocorrelation peak (Fig. 2c) as it is the case in real OCT signals, the predicted dispersion profile (Fig. 2d, orange line) is affected in the areas around that peak (marked as 0 next the gray line representing an A-scan in Fig. 2d). This is because the neural network, which was trained on autocorrelation-peak-free signals, "treats" the autocorrelation peak as a structural peak. Consequently, as one would expect, GVD in front of the auto-correlation peak is the same as GVD of the object layer. Since the autocorrelation peak comes from the interference of light back-scattered from the first and the second interface of the object, it contains uncompensated dispersion equal to the dispersion of the object layer. Interestingly, the part of the dispersion profile between the autocorrelation peak and the first interface peak (peaks 0 and 1 in the A-scan in Fig. 2d) shows a non-zero GVD value, although that area represents zero-GVD air.

To find out why some places in the dispersion profiles show false GVD levels, theoretical calculations reported in the Supplementary Document were carried out. Equations describing a signal in two situations were derived: one for the case where the object consists of two interfaces with a signal incorporating an autocorrelation peak and one for the case of a three-interface object with a signal with no autocorrelation peaks. Whereas the former case represents the data displayed in Fig. 2c and general experimental signals, the latter one represents the data used for neural network training. Observing that the first peak in a three-interface object corresponds to the autocorrelation peak in the two-interface object A-scan, a formula is found for GVD of the layer between the autocorrelation peak and the first interface, $\beta_{NL,2}^{(even)}$, as a function of the GVD of the front layer (air in this example), $\beta_{NL,front}^{(even)}$, and the object layer, $\beta_{NL,obj}^{(even)}$:

$$\beta_{NL,2}^{(even)} = \frac{L_{front}\beta_{NL,front}^{(even)} - L_{obj}\beta_{NL,obj}^{(even)}}{L_{front} - L_{obj}} \tag{1}$$

where $L_{front}$ is the distance between 0 OPD point and the first object interface, and $L_{obj}$ is the object thickness. Substituting $L_{front}$ = 360, $L_{obj}$ = 260 and $\beta_{NL,front}^{(even)}$ = 0 fs$^2$/mm gives $\beta_{NL,2}^{(even)}$ = –2,000 fs$^2$/mm which matches the GVD level in the predicted dispersion profile. This result confirms that indeed the neural network interprets the auto-correlation peak as a structural peak and consequently, outputs GVD levels corresponding to a three-interface object.

In Fig. 2c, the autocorrelation peak was positioned in front of the layer. In Fig. 3c, we simulated an FFT stack for an object for which the autocorrelation peak is found between the structural peaks. The object simulated in Fig. 3 has a layer with GVD equal to 2,000 fs$^2$/mm. Also, the area in front of the layer has non-zero GVD equal to 1,000 fs$^2$/mm, which means that there is dispersion imbalance in the interferometer due to, for example, excess amount of glass in the reference arm. Whereas in the absence of the autocorrelation peak (Fig. 3a) the neural network prediction gives a good estimation of the GVD value distribution (orange line in Fig. 3b), the prediction based on the FFT stack incorporating an autocorrelation peak shows discrepancies again (Fig. 3d).

In the case of the autocorrelation peak situated between the object peaks, it is the second peak in the three-interface object example which corresponds to the autocorrelation peak. In such a case, the following relationships are found for the GVD level between the first interface peak and the autocorrelation peak, $\beta_{NL,2}^{(even)}$, and the GVD level between the autocorrelation peak and the second interface peak, $\beta_{NL,3}^{(even)}$:

$$\beta_{NL,2}^{(even)} = \frac{L_{front}\beta_{NL,front}^{(even)} - L_{obj}\beta_{NL,obj}^{(even)}}{L_{front} - L_{obj}} \tag{2a}$$

$$\beta_{NL,3}^{(even)} = \beta_{NL,front}^{(even)} \tag{2b}$$

In the case depicted in Fig.3c, $L_{front}$ = 200 and $L_{obj}$ = 350, which gives $\beta_{NL,2}^{(even)}$ = 2667 fs$^2$/mm and $\beta_{NL,3}^{(even)}$ = 1000 fs$^2$/mm. Both of the calculated dispersion values match the ones predicted by the neural network.

In summary, our theoretical analysis together with the simulations show that the neural network effectively treats a two-interface object with an auto-correlation peak as a three-interface object without any autocorrelation. Similarly, a three-interface object with auto-correlation peaks will be treated by the neural network as a six interface object (three structural peaks plus three autocorrelation peaks, each corresponding to a pair of structural interfaces), four-interface object with autocorrelation peaks will be treated as a ten-interface object (four structural peaks plus six auto-correlation peaks), and so on. In general, theoretically, an $N$-interface object with autocorrelation peaks will be treated as an $N(N+1)/2$-interface object without autocorrelation.

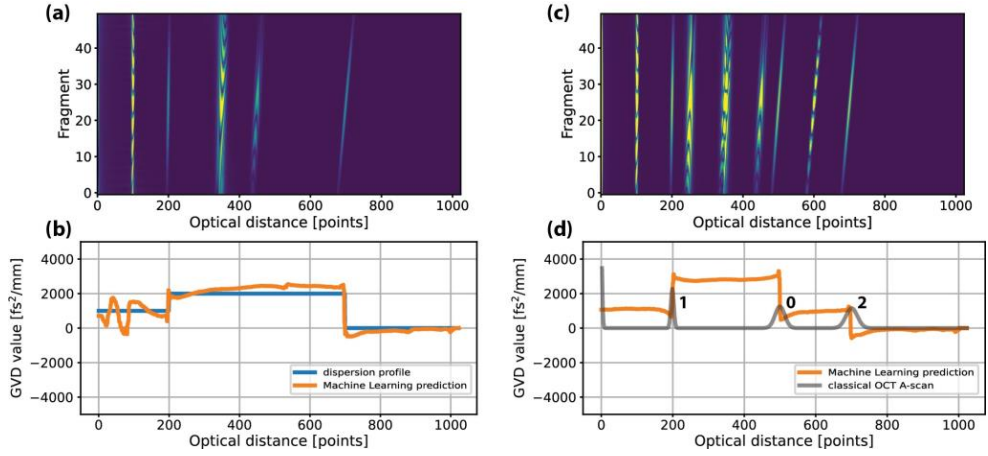

Fig. 3. A two-interface object with object dispersion equal to 2,000 fs$^2$/mm and uncompensated interferometer dispersion of 1,000 fs$^2$/mm. (a) FFT stack - an input of the neural network - does not contain an autocorrelation peak. (b) The neural network prediction - orange line - is in good agreement with the ground truth - blue line, except for the parts close to 0 OPD. (c) When the FFT stack contains the autocorrelation peak, (d) the neural network prediction - orange line - shows a dispersion profile affected around the location of the autocorrelation peak (marked as 0 on an A-scan plotted in gray), which is positioned between the interface peaks (marked as 1 and 2).

*Dispersion profiles - computer-generated data with noise*

We have checked how the quality of predictions change with different levels of noise. A signal for a two-interface object, the same as the one from Fig. 3, was created with six different signal-to-noise ratio (SNR) values: 70 dB (Fig. 4a), 75 dB (Fig. 4b), 80 dB (Fig. 4c), 83 dB (Fig. 4d), 86 dB (Fig. 4e) and 90 dB (Fig. 4f). Fig. 4g-l show the neural network predictions (orange line) and the ground-truth dispersion profile where the changes due to the presence of the autocorrelation peak are accounted for (blue line).

It is observed that for the lower SNR values the neural network is not able to correctly estimate the GVD values, producing substantial mistakes for signals with SNR lower than 90 dB. This is due to the fact that the noise is treated as a structure and as a result, the network outputs GVD values corresponding to the elements appearing in an FFT stack due to the noise. The majority of the falsely predicted GVD values are equal to either 5,000 or −5000 fs$^2$/mm, which are the upper and lower limits of the GVD value range used during the training. The prevalence of these two values might result from the fact that the noise-related elements in the FFT stack are "seen" by the neural network as very high or very low dispersion structural elements. Since there is an upper and lower limit with which the network was trained, the very high or very low GVD values are outputted as 5,000 or −5000 fs$^2$/mm.

*Dispersion profiles - experimental data*

Three pieces of glass were used to test the performance of the neural network in the experimental conditions: 50-$\mu$m thick quartz, 1000-$\mu$m thick BK7 and 750-$\mu$m thick sapphire. The predictions (in orange) together with A-scans (in light gray) are depicted in Fig. 5a-c. In the case of BK7 and sapphire, one can see the influence of the autocorrelation peak in the dispersion profiles. The prediction for quartz seems to incorporate many errors, most probably due to a higher level of noise than in the other cases, which is why it will not be further discussed. However, quick calculations can be made for BK7 and sapphire to estimate their GVD.

For BK7, the distance between 0 OPD point and the first interface peak (marked with 1 in Fig. 5b) is around 220 points (=$L_{front}$), the distance between the first and the second interface peaks (1 and 2 in Fig. 5b) is around 700 points (=$L_{obj}$). Also, as expected, the GVD level between 0 OPD point and the first interface peak is similar to the GVD level between the autocorrelation peak (marked with 0 in Fig. 5b) and the second interface peak and is equal to around 2000 fs$^2$/mm (=$\beta_{NL,front}^{(even)}$). Finally, the GVD level between the first interface peak and the autocorrelation peak is around −850 fs$^2$/mm (= $\beta_{NL,2}^{(even)}$). Using (2) gives $\beta_{NL,obj}^{(even)} \approx 46$ fs$^2$/mm. The literature GVD value for BK7 at 840 nm is 41 fs$^2$/mm. Similarly, for sapphire, $L_{front} \approx 70$, $L_{obj} \approx 260$, $\beta_{NL,front}^{(even)} \approx 3700$ fs$^2$/mm, and $\beta_{NL,2}^{(even)} \approx -1300$ fs$^2$/mm. Using (2) gives $\beta_{NL,obj}^{(even)} \approx 46$ fs$^2$/mm. The literature GVD value for sapphire at 840 nm is 53 fs$^2$/mm.

It should be noted that the estimation of GVD values using predicted dispersion profiles is very rough and intrinsically burdened with a big error. This is due to the fact that the dispersion levels predicted by the neural network are highly variable and therefore, provide wide ranges of possible GVD values.

Also, as it can be seen in Fig. 5a-c, the locations of the layers in the dispersion profiles do not overlap with the locations of the peaks in the A-scans. This is caused by the fact that the uncompensated interferometer dispersion, apart from broadening each peak in the A-scan, leads to their displacement by a constant distance.

The sapphire glass was laterally scanned to obtain a two-dimensional image called a B-scan (Fig. 5e). The B-scan shows the structure of the sapphire in the form of two tilted lines (the sapphire was at a small angle during scanning) as well as autocorrelation peaks which form a vertical line. Each spectrum behind the A-scans that comprise the B-scan is taken to be processed and transformed with the neural network, resulting in dispersion profiles which - when put one on top of another - create a dispersion map (Fig. 5d). Such a dispersion map clearly shows where each layer is and what level of GVD it is characterised with.

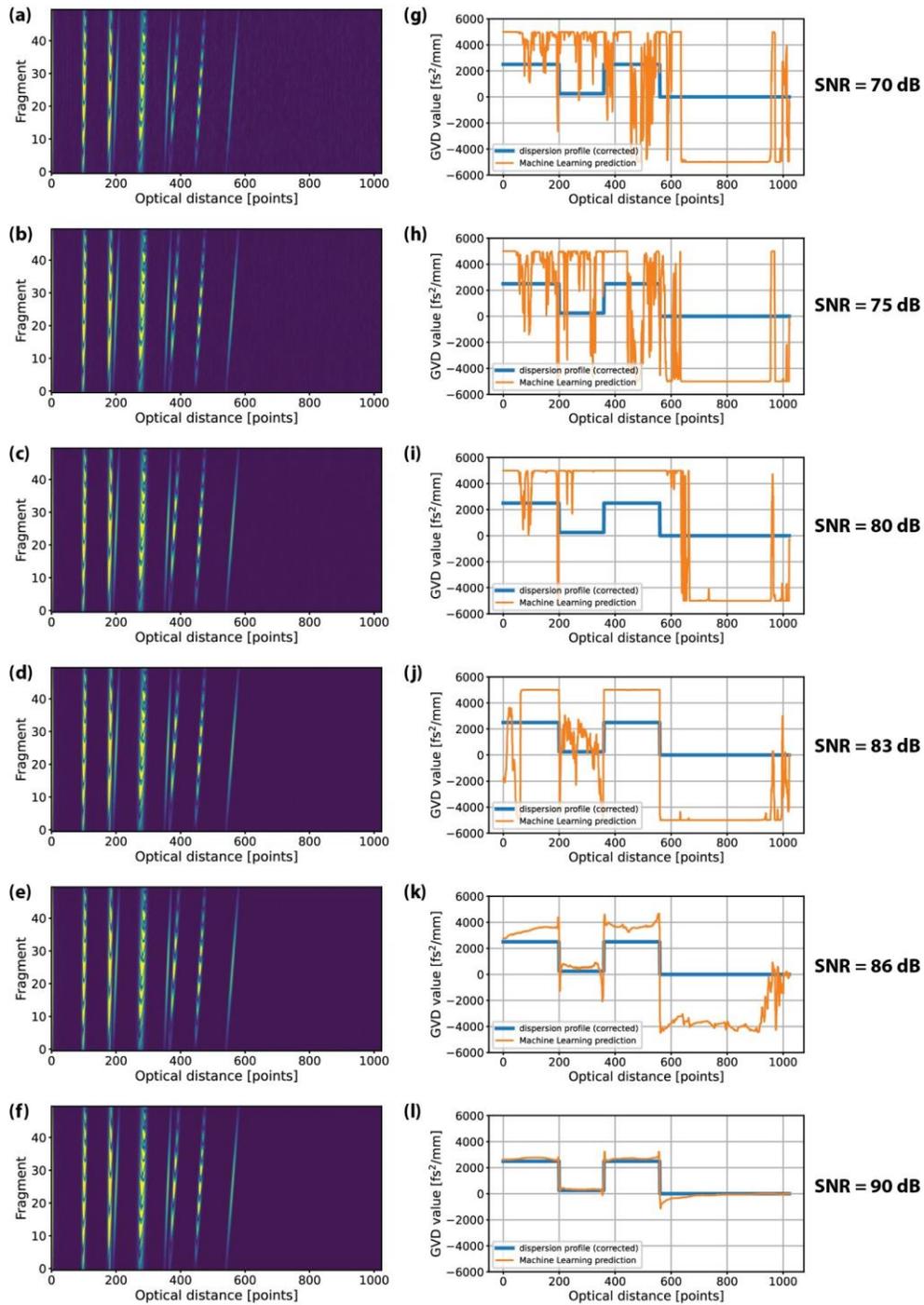

Fig. 4. (a-f) FFT stacks and (g-l) the corresponding neural network predictions (orange line) and ground-truth dispersion profiles (blue line) for six different SNR.

The autocorrelation peaks were removed from A-scans in the sapphire B-scan (Fig. 5g) and inverse Fourier transformed to spectra. Such filtered spectral data was processed with the neural network and the dispersion map (Fig. 5f) was obtained. Again, there's a high positive GVD level in the area between 0 OPD and the first interface, suggesting a dispersion imbalance in the interferometer. Due to the lack of the autocorrelation peaks, the area between the interface peaks does not show the rapid GVD level change observed in Fig. 5d, but it is not uniform, either. Although it should be a constant level of 53 fs$^2$/mm throughout the area of the object layer, it incorporates fluctuations which most probably appear due to the noise. After the removal of the autocorrelation peaks, the GVD level of the object layer dropped to the height comparable to the actual GVD of sapphire with the exception of some regions for which GVD remained intact (for example the negative-GVD area in the lower left corner). The presence of these regions indicates an existence of low-intensity peaks between the interface peaks.

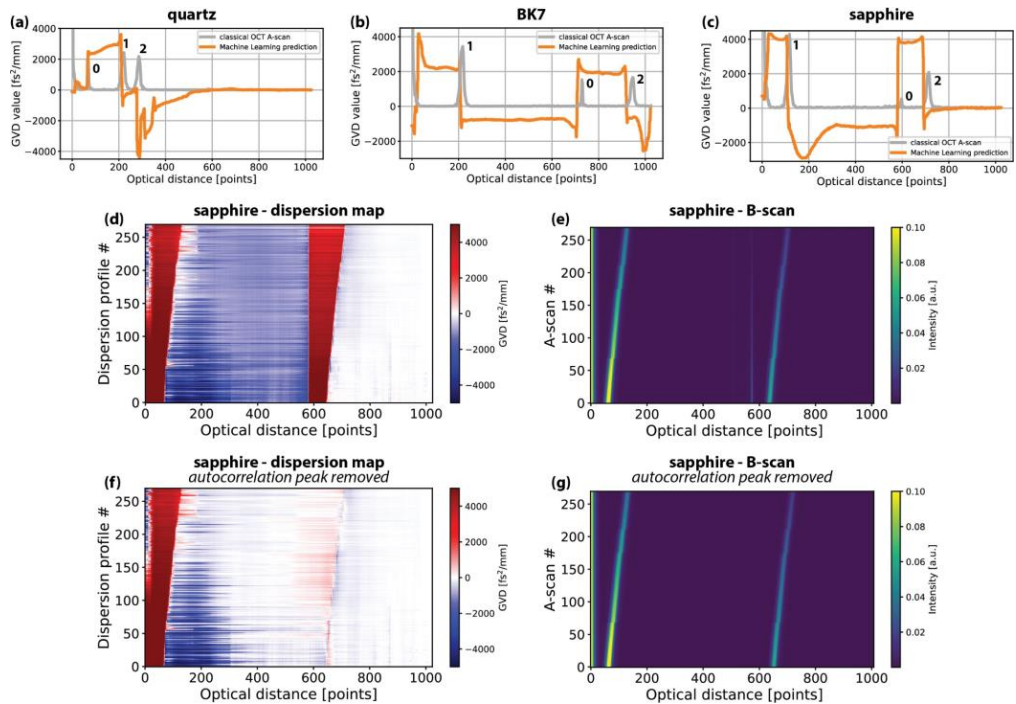

Fig. 5. Neural network predictions (orange line) together with corresponding A-scans (gray line) for (a) quartz, (b) BK7, and (c) sapphire. (d) A dispersion map created using data behind A-scans in (e) a B-scan of a sapphire glass put at a slight angle while scanning. (f) A dispersion map for the same sapphire after the autocorrelation peaks were removed from each A-scan, as depicted on (g) the B-scan. 0 - autocorrelation peak, 1 and 2 - object interface peaks.

**Discussion and future work**

Quantum-mimic OCT signal is combined with Machine Learning to provide estimations of GVD value distribution within the imaged object. This approach is tested on simple two-interface objects for which the signals were either computer-generated or acquired with an OCT system. Since the neural network was trained on synthesised signals which did not incorporate autocorrelation peaks, the predictions for real-life-like signals, i.e. signals containing autocorrelation peaks, showed shifts in GVD levels around the location of the autocorrelation peaks. The extent of these shifts is used to calculate an estimate of GVD for two objects: BK7 and sapphire. Because the predicted GVD levels are highly variable in their height, these estimates are burdened with a large error. Consequently, the dispersion profiles provided by the neural network are more of a qualitative nature.

Also, autocorrelation peaks are inherent to OCT imaging and if not removed using experimental or algorithmic means, will affect the real GVD levels. The experimental removal of autocorrelation peaks is performed by putting an object at an angle to the direction of light propagation. The algorithmic approach, as it was described in the previous section) is based on zeroing the elements at the position of the autocorrelation peaks and inverse Fourier transforming such A-scan back to the spectrum. To remove the detrimental influence of autocorrelation peaks, their presence could be accounted for in the signals used for training. However, as was shown earlier [10] where such training was performed, a small deviation of an FFT stack from its ideal representation - mainly introduced by noise - leads to GVD level shifts similar to those observed in the case of training performed on data which do not account for autocorrelation peaks.

The results presented in this article confirm that a quantum-mimic OCT signal contains enough information about layer-specific dispersion in the imaged object and a neural network could be successfully trained to retrieve it. Future work will consist in optimising the neural network parameters to obtain more precise predictions and consequently, results providing quantitative information on GVD.

**Acknowledgements**
P.K. would like to acknowledge financial support by the Foundation for Polish Science (FNP) (project First Team co-financed by the European Union under the European Regional Development Fund, grant no. First Team/2017-3/20) and National Laboratory of Atomic, Molecular and Optical Physics in Torun, Poland. V.V. and K.A.M would like to thank Heyang (Thomas) Li and Oliver Batchelor for providing additional GPU resources.

**Disclosures**

The authors declare no conflicts of interest.

See Supplement 1 for supporting content.